%% file: main.tex
\documentclass[10pt,twocolumn,letterpaper]{article}
\PassOptionsToPackage{inline}{enumitem}
\usepackage{cvpr}              % To produce the CAMERA-READY version
\input{preamble}

\definecolor{cvprblue}{rgb}{0.21,0.49,0.74}

\usepackage{graphicx}
\usepackage{amsmath}
\usepackage{amssymb}
\usepackage{booktabs}% for \midrule and \cmidrule macros
\usepackage{multirow}% for \multirow
\usepackage{colortbl}% for \cellcolor
\usepackage{breqn}% for automatic linebreaks in long equations
\usepackage{accents}% for doublehat, defined in macros
\usepackage{adjustbox}% for adjustbox for large table
\usepackage{diagbox} % for diagbox in large table
\usepackage{xcolor} % for colored text, e.g., when refering to a colored point in a figure
\usepackage[pagebackref,breaklinks,colorlinks,citecolor=cvprblue]{hyperref}

\usepackage[accsupp]{axessibility}

\def\paperID{4083} % *** Enter the Paper ID here
\def\confName{CVPR}
\def\confYear{2024}

\title{A Call to Reflect on Evaluation Practices for Age Estimation:\\Comparative Analysis of the State-of-the-Art and a Unified Benchmark}

\author{Jakub Paplh\'am\\
Department of Cybernetics\\ Faculty of Electrical Engineering\\
Czech Technical University in Prague\\
{\tt\small paplhjak@fel.cvut.cz}
\and
Vojt\v ech Franc\\
Department of Cybernetics\\ Faculty of Electrical Engineering\\
Czech Technical University in Prague\\
{\tt\small xfrancv@cmp.felk.cvut.cz}
}

\begin{document}
\maketitle

%%%%%%%%% ABSTRACT
\begin{abstract}
Comparing different age estimation methods poses a challenge due to the unreliability of published results stemming from inconsistencies in the benchmarking process. Previous studies have reported continuous performance improvements over the past decade using specialized methods; however, our findings challenge these claims. This paper identifies two trivial, yet persistent issues with the currently used evaluation protocol and describes how to resolve them. %We describe our evaluation protocol in detail and provide specific examples of how the protocol should be used. 
We %utilize the protocol to 
offer an extensive comparative analysis for state-of-the-art facial age estimation methods. Surprisingly, we find that the performance differences between the methods are negligible compared to the effect of other factors, such as facial alignment, facial coverage, image resolution, model architecture, or the amount of data used for pretraining. We use the gained insights to propose using FaRL as the backbone model and demonstrate its effectiveness on all public datasets. We make the source code and exact data splits public on \href{https://github.com/paplhjak/Facial-Age-Estimation-Benchmark}{GitHub} and in the supplementary material. %The source code is available at .
\end{abstract}

%%%%%%%%% BODY TEXT
\section{Introduction}
\label{sec:intro}
Age estimation has received significant interest in recent years. However, a closer examination of the evaluation process reveals two underlying issues. First, no standardized data splits are defined for most public datasets, and the used splits are rarely made public, making the results irreproducible. Second, methods often modify multiple components of the age estimation system, making it unclear which modification is responsible for the performance gains.

This paper aims to critically analyze the evaluation practices in age estimation research, highlight the issues, and appeal to the community to follow good evaluation practices to resolve them. We benchmark and fairly compare recent deep-learning methods for age estimation from facial images. We focus on \textit{state-of-the-art} methods that adapt a generic architecture by changing its last layer or the loss function to suit the age estimation task. Although this may appear restrictive, it is essential to note that most of the methods proposed in the field fall into this category ($\approx 70\%$). By comparing methods that modify only a small part of the network, we aim to ensure a fair evaluation, as the remaining setup can be kept identical. Besides the usual intra-class performance, we also evaluate their cross-dataset generalization, which has been neglected in the age prediction literature so far. Surprisingly, we find that the influence of the loss function and the decision layer on the results, usually the primary component that distinguishes different methods, is negligible compared to other factors.

\paragraph{Contributions}
\begin{itemize}
\item We show that existing evaluation practices in age estimation do not provide consistent results. This leads to obstacles for researchers aiming to advance prior work and for practitioners striving to pinpoint the most effective approach for their application.

\item We define a proper evaluation protocol, offer an extensive comparative analysis for state-of-the-art facial age estimation methods, and publish our code.

\item We show that the performance difference caused by using a different decision layer or training loss is significantly smaller than that caused by other parts of the prediction pipeline. 

\item We identify that the amount of data used for pre-training is the most influential factor and use the observation to propose using FaRL \cite{FaRL} as the backbone architecture. We demonstrate its effectiveness on public datasets.
%We fairly benchmark existing state-of-the-art approaches and show that differences in their performance are negligible.
\end{itemize}

\begin{figure}[htb]
  \centering\includegraphics{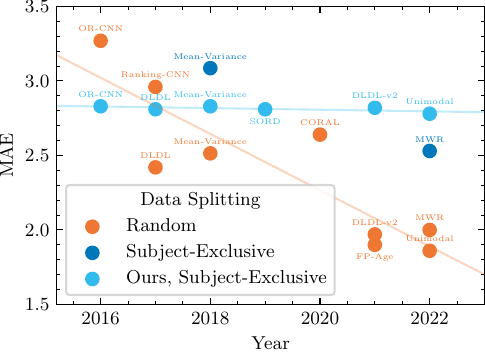}
   \caption{Mean Absolute Error (MAE) $\downarrow$ of age estimation methods on the MORPH dataset, as reported in the existing literature and measured by us, viewed over time. {\color{colorq1}Random splitting} remains the prevalent data splitting strategy. The consistent performance improvements over time are attributed in the literature to specialized loss functions for age estimation. {\color{colorq2}Subject-exclusive} (identity-disjoint) data splitting is rarely employed. With {\color{colorq3}unified subject-exclusive} data splitting and all factors except the loss function fixed, all evaluated methods yield comparable results, failing to achieve the performance gains promised by the random splitting.}   
\label{fig:year_vs_mae}
\end{figure}

\section{Issues with Current Evaluation Practices}
\subsection{Data Splits}
Publications focused on age estimation evaluate their methods on several datasets \cite{AgeDB, CLAP, CACD, MORPH, UTKFace, IMDB, OR-CNN}. The most commonly used of these is the MORPH \cite{MORPH} dataset. However, the evaluation procedures between the publications are not unified. For instance, OR-CNN \cite{OR-CNN} \textit{randomly} divides the dataset into two parts: 80\% for training and 20\% for testing. No mention is made of a validation set for model selection. \textit{Random splitting} (RS) protocol is also used in \cite{FP-AGE, MegaAge, CORAL, DLDL, DLDL-V2, RankingCNN}, but the specific \textit{data splits} differ between studies as they are rarely made public. Since the dataset contains multiple images per person (many captured at the same age), the same individual can be present in both the training and testing sets. This overlap introduces a bias, resulting in overly optimistic evaluation outcomes. The degree of data leakage can vary when using random splitting, making certain data splits more challenging than others. Further, this fundamentally changes the entire setup; rarely will one want to deploy the age estimation system on the people present in the training data. Consequently, comparison of different methods and discerning which method stands out as the most effective based on the published results becomes problematic. 

Only some publications \cite{MeanVariance, MovingWindowRegression} recognize this bias introduced by the splitting strategy and address it by implementing \textit{subject-exclusive} (SE) \cite{MeanVariance} splitting. This approach ensures that all images of an individual are exclusively in the \begin{enumerate*}[label=(\roman*),before=\unskip{ }, itemjoin={{, }}, itemjoin*={{, or }}] 
\item training
\item validation
\item testing part.
\end{enumerate*} The terminology here is not fully established. One might encounter either \textit{identity-disjoint} or \textit{person-disjoint} instead of subject-exclusive in the literature.

To assess how prevalent \textit{random splitting} (RS) on the MORPH dataset truly is, we conducted a survey of all age estimation papers presented at the %Computer Vision and Pattern Recognition Conference 
CVPR and %the International Conference on Computer Vision (
ICCV since 2013. We found 16 papers focused on age estimation, of which nine use RS, two use SE, five use specialized splits, and three do not utilize MORPH. We further surveyed other research conferences and journals, namely: %International Joint Conference on Artificial Intelligence (
IJCAI, %British Machine Vision Conference (
BMVC, %Asian Conference on Computer Vision 
ACCV, %IEEE Transactions on Image Processing
IEEE TIP, Pattern Recognit. Lett., Pattern Anal. Appl., and find eight influential age estimation papers that use MORPH. Of those, seven use RS, and one uses a specialized split. By specialized splits, we are referring to non-standard strategies such as ethnically balanced partitions.

Altogether, we discover that only $\approx 10\%$ of papers that utilize MORPH use the SE protocol. This finding is concerning, as MORPH \cite{MORPH} is the most popular dataset used to compare age estimation approaches. Other datasets do not provide a reliable benchmark either, as standardized data splits are provided only for two public age estimation datasets: \begin{enumerate*}[label=(\roman*),before=\unskip{ }, itemjoin={{, }}, itemjoin*={{, and }}]
    \item the ChaLearn Looking at People Challenge 2016 (CLAP2016) dataset \cite{CLAP}, which is relatively small, consisting of fewer than 8000 images
    \item the Cross-Age Celebrity Dataset (CACD2000) \cite{CACD}, which has noisy training annotations and is not intended for age estimation. 
    \end{enumerate*}
    Comparing methods using only these datasets is, therefore, not satisfactory either. Other popular datasets, AgeDB dataset \cite{AgeDB} and Asian Face Age Dataset (AFAD) \cite{OR-CNN}, also consist of multiple images per person, requiring SE splitting. However, they lack any data splits accepted by the community and often are used with the RS protocol. As such, they suffer from the same issues as MORPH \cite{MORPH}.

%Surprisingly, only Pan \etal \cite{MeanVariance} and \mbox{Shin \etal \cite{MovingWindowRegression}} utilized subject-exclusive splitting, indicating that the majority of studies rely on random splitting as their methodology.

\subsection{Pipeline Ablation}
\label{ssec:pipeline_components}
To fairly compare multiple methods, an identical experimental setup should be used for each of them. The current state-of-the-art age estimation approaches adhere to a common framework encompassing:
\begin{enumerate*}[label=(\roman*),before=\unskip{ }, itemjoin={{, }}, itemjoin*={{, and }}]
    \item data collection% the specific data for training the model and evaluating its performance.
    \item data preprocessing%: This involves preparing the raw data to create inputs for the neural network.
    \item model design, including the decision layer and the loss function
    \item training and evaluation.
\end{enumerate*}
Most novel approaches introduce distinct changes to the component (iii); namely they design a specialized loss function to exploit the ordinal nature of age. However, they frequently alter multiple components of the framework simultaneously, complicating the attribution of performance improvements to the claimed modifications.

To compare different loss functions, e.g., \cite{OR-CNN, CORAL, MeanVariance, Unimodal, EBC, DLDL, DLDL-V2, SoftLabels}, the other components of the framework should be kept constant, allowing us to isolate the impact of the selected method on the performance. This is trivial, yet the age estimation community mostly ignores it. Further, many publications hand-wave the other components and do not precisely specify them, making future comparisons meaningless. {\em It's important to question whether the reported enhancement in a research paper truly stems from the novel loss function it proposes or if it could be attributed to a different modification.} We strongly advocate that each component be addressed in isolation and that the experimental setup be precisely described.

Over the past decade, numerous novel age estimation methods have been introduced, promising continuous performance improvements every year. However, motivated by these findings, we raise the question: how reliable are the published age estimation results? In \cref{sec:evaluation_protocol} we aim to establish a proper evaluation protocol and use it in \cref{sec:comparative_method_analysis} to compare the methods \cite{DEX, DLDL, DLDL-V2, MeanVariance, Unimodal, SoftLabels, OR-CNN} reliably. \Cref{fig:year_vs_mae} illustrates the contrast between the performance of state-of-the-art methods as reported in their respective studies and the outcomes as measured by our implementation.%achieved through the implementation of our proposed evaluation protocol.

\section{Evaluation Protocol}
\label{sec:evaluation_protocol}

\input{vf_proposal_eval_protocol.tex}

\section{Comparative Method Analysis}
\label{sec:comparative_method_analysis}
This section applies the evaluation protocol to compare state-of-the-art age estimation methods. We maintain a consistent preprocessing procedure, model architecture, and dataset while selectively altering the decision layer and loss function to incorporate modifications proposed in prominent works such as \cite{DEX, DLDL, DLDL-V2, MeanVariance, Unimodal, SoftLabels, OR-CNN}.
%, In order to replicate a setup for age estimation, using the same data split, is insufficient. Unlike other deep learning tasks, where the entire image is often used as input, age estimation models should only be provided with a specific region of the image that corresponds to the face. The selection of this region then undoubtedly influences the model's performance. To faithfully reproduce the setup, the whole pipeline, therefore, needs to be reproduced. 
%The main challenge in comparing different age estimation methods lies in the fact that published results often utilize different data splits. Even when using a consistent data-splitting protocol, such as random or subject-exclusive, the results can differ significantly (i.e., the difference is on the same level as the impact of a different loss function) if the specific data splits are not identical. To address this, we evaluate all methods on identical data splits. 
%In this section, we discuss the used datasets and the data-splitting strategy.
\subsection{Methodology}
 \paragraph{Datasets} We evaluate the methods using 7 datasets: AgeDB \cite{AgeDB}, AFAD \cite{OR-CNN}, CACD2000 \cite{CACD}, CLAP2016 \cite{CLAP}, FG-NET \cite{FG-NET}, MORPH \cite{MORPH}, and UTKFace \cite{UTKFace}. We also use the IMDB-WIKI dataset \cite{IMDB} for pre-training with clean labels from Franc and \v Cech \cite{EM-CNN}. 

 \paragraph{Data Splits}
For the CLAP2016 and CACD2000 datasets, we use the single data split provided by the dataset authors. For the remaining datasets, we create five subject-exclusive (SE) data splits. To generate the split, we partition the dataset such that $60\%$ of the dataset is used for training, $20\%$ for model selection (validation), and $20\%$ for evaluating the model performance (test). Additionally, we ensure that each partition has the same age distribution. Due to its small size, we only use FG-NET for evaluation. We make our data splits and code public at \href{https://github.com/paplhjak/Facial-Age-Estimation-Benchmark}{Facial-Age-Benchmark}\footnote{https://github.com/paplhjak/Facial-Age-Estimation-Benchmark}.

\paragraph{Model Architecture \& Weight Initialization}
We use ResNet-50 \cite{ResNet} as the backbone architecture. We always start the training of the methods from the same initialization. We run the experiments with \begin{enumerate*}[label=(\roman*),before=\unskip{ }, itemjoin={{, }}, itemjoin*={{, and }}]
\item random initialization
\item weights pre-trained on ImageNet (TorchVision's \textup{IMAGENET1K\_V2})
\item weights pre-trained on ImageNet and then further trained on IMDB-WIKI for age estimation with cross-entropy.
\end{enumerate*}
After the pre-training, the last layer of the model is replaced with a layer specific to the desired method. The models are then fine-tuned on the downstream dataset. It is important to note that for the baseline cross-entropy, we also replace the final layer before fine-tuning. This ensures that the experimental setup remains identical to that of the other methods.

\paragraph{Training Details}
We utilize the Adam optimizer with the parameters $\beta_1=0.9$, $\beta_2=0.999$. For pre-training on the IMDB-WIKI dataset, we set the learning rate to ${\alpha=10^{-3}}$ and train the model for a total of 100 epochs. For fine-tuning on the remaining datasets we reduce the learning rate to ${\alpha=10^{-4}}$ and train the model for 50 epochs. We use a batch size of 100. The best model is selected based on the MAE metric computed on the validation set. We utilize two data augmentations during training, \begin{enumerate*}[label=(\roman*),before=\unskip{ }, itemjoin={{, }}, itemjoin*={{, and }}]
    \item horizontal mirroring
    \item cropping out an 80\% to 100\% portion of the bounding box and resizing it to the model input shape.
\end{enumerate*} 
We do \textit{not} tune the hyperparameters of the methods \cite{OR-CNN, DLDL, DLDL-V2, SoftLabels, MeanVariance, Unimodal} on the validation set. We apply them in the original configurations. We argue that if any of the loss functions is a significant improvement over the baseline, we should observe a performance improvement across a broad range of hyperparameters and preprocessing pipelines. We consider our training parameters to be reasonable and to provide a comparison of the methods as if employed \textit{out-of-the-box}.

\paragraph{Preprocessing}
We use the RetinaFace model developed by Deng \etal \cite{RetinaFace} for face detection and facial landmark detection. We use complete \textit{facial coverage}, i.e., the images encompass the entire head. We resize the images to a resolution of $256\times256$ pixels and normalize the pixel values of the images. To this end, we subtract the mean and divide by the standard deviation of colors on ImageNet \cite{ImageNet}.

\paragraph{Metrics}
We use the Mean Absolute Error (MAE) calculated on the test data as the performance measure. 
To determine whether any method is consistently better than others, we employ the Friedman test and the Nemenyi critical difference test (FN test) as described by Dem{\v{s}}ar \cite{demvsar}. The main statistic used in the test is the average ranking ($1$ is best) of a method computed on multiple datasets. Differences in the average ranking are then used to decide whether a method is significantly better than others or whether the improvement is due to randomness (the null hypothesis). We use a common significance level (p-value) of $\alpha=5\%$.

\subsection{Results}
\paragraph{Intra-Dataset Performance}
The intra-dataset results can be seen in \cref{tab:large_table}, highlighted with a grey background. When starting from random initialization, training with the \mbox{Unimodal loss \cite{Unimodal}} tends to be unstable. Excluding the Unimodal loss \cite{Unimodal} from the evaluation, we apply the FN test. The results indicate that three methods: OR-CNN \cite{OR-CNN}, DLDL \cite{DLDL}, and the Mean-Variance loss \cite{MeanVariance}, demonstrate a significant performance improvement over the baseline cross-entropy. \textit{With limited data availability, when pre-training is not possible, it is advisable to utilize one of the aforementioned methods.} 

With pre-training, either on ImageNet or IMDB-WIKI, none of the methods is significantly better than the cross-entropy. In other words, \textit{we do not observe any systematic improvement by deviating from the standard approach}. 

%With the IMDB-WIKI pre-training, the average ranking ($1$ is best) of each method for intra-dataset performance is \begin{enumerate*}[label=(\roman*),before=\unskip{ }, itemjoin={{, }}, itemjoin*={{, and }}]
%    \item Cross-Entropy: $3.50$
%    \item OR-CNN \cite{OR-CNN}: $3.83$
%    \item DLDL \cite{DLDL}: $\mathbf{2.67}$
%    \item DLDL-v2 \cite{DLDL-V2}: $3.92$
%    \item SORD \cite{SoftLabels}: $\underline{3.00}$
%    \item Mean-Variance \cite{MeanVariance}: $5.92$
%    \item Unimodal \cite{Unimodal}: $5.17$
%\end{enumerate*}.

 \paragraph{Cross-Dataset Generalization}
Cross-dataset results, shown in \cref{tab:large_table} with white background, were obtained by evaluating the performance of models on datasets that were not used for their training. The cross-dataset performance is unsurprisingly significantly worse than the intra-dataset performance for all of the methods. Using the FN test, we conclude that there is \textit{no significant difference} in generalization capability between any of the methods \cite{OR-CNN, DLDL, DLDL-V2, SoftLabels, MeanVariance, Unimodal} and the cross-entropy, regardless of pre-training. In other words, though the loss functions may reduce overfitting, they do not help in the presence of covariate shift.

%With the IMDB-WIKI pre-training, the average ranking ($1$ is best) of each method for cross-dataset performance is \begin{enumerate*}[label=(\roman*),before=\unskip{ }, itemjoin={{, }}, itemjoin*={{, and }}]
%    \item Cross-Entropy: $4.42$
%    \item OR-CNN \cite{OR-CNN}: $\mathbf{3.35}$
%    \item DLDL \cite{DLDL}: $3.62$
%    \item DLDL-v2 \cite{DLDL-V2}: $3.93$
%    \item SORD \cite{SoftLabels}: $\underline{3.40}$
%    \item Mean-Variance \cite{MeanVariance}: $4.47$
%    \item Unimodal \cite{Unimodal}: $4.81$
%\end{enumerate*}.

None of the methods perform well when evaluated on a different dataset than the one they were trained on. The best cross-dataset results are achieved by training on either UTKFace or CLAP2016. The worst performance across databases is observed when models are trained on AFAD or MORPH. This discrepancy can be attributed to UTKFace and CLAP2016 having a broader range of images, which allows them to generalize effectively to other datasets. Conversely, the limited diversity in MORPH or AFAD datasets, such as AFAD mainly comprising images of people of Asian ethnicity and around $80\%$ of MORPH being composed of individuals of African American ethnicity, contributes to the poor knowledge transfer. The significant decrease in the performance of models trained on the MORPH dataset when applied to other age estimation datasets underscores the importance of not relying solely on the MORPH dataset as the benchmark for age estimation. To ensure a reliable evaluation of different methods, it is crucial to incorporate results from alternative datasets as well.

%\paragraph{Pre-training Loss Function}
%It can be argued that the utilization of cross-entropy for the IMDB-WIKI pre-training puts the other methods at a disadvantage, as the extracted features become more suitable for cross-entropy rather than for the alternative methods. In order to investigate this possibility, we have chosen the two most recent methods, namely the Mean-Variance loss proposed by Pan \etal \cite{MeanVariance} and the Unimodal loss presented by Li \etal \cite{Unimodal}. We pre-train the models on IMDB-WIKI using the corresponding loss functions and subsequently finetune them on the downstream datasets. The obtained results are displayed in \cref{tab:intra_dataset_with_native_pre-training}. We do not find any benefit when using the same method for both pre-training and finetuning. On the contrary, we observe that the Unimodal loss \cite{Unimodal} yields better results across all datasets when the model is pre-trained using cross-entropy. Thereforwe conclude that the cross-entropy pre-training does not invalidate our results.

\section{Component Analysis}
In this section, we analyze the influence of the backbone architecture and the data preparation pipeline on model performance. We show that changes to these components can have a much more significant impact on the final performance than the choice of a loss function. When altering a component, we maintain all other components at their defaults, presented as the Cross-Entropy approach in \cref{sec:comparative_method_analysis}. We use the gained insight to propose a strong baseline age estimation model using the FaRL \cite{FaRL} backbone.

\subsection{Model Architecture}
\label{ssec:architecture}
Multiple different backbone architectures can be found in the age estimation literature. Among these architectures, VGG16 \cite{DLDL-V2, MegaAge, MovingWindowRegression, MeanVariance, Unimodal, SoftLabels} and ResNet-50 \cite{CORAL, LabelDiversity, FP-AGE} stand out as the most common choice. We evaluate the influence of the architecture choice on the performance and extend the comparison to include more recent advancements, EfficientNet-B4 and ViT-B-16. We present our findings in \cref{tab:model_architecture}. No backbone emerges as universally best across all datasets. Notably, changes in the backbone have a more substantial impact on performance than changes to the loss function, see \cref{tab:large_table}. This highlights the importance of a thorough ablation, as changes in the backbone architecture could obscure the impact of the loss function.

\subsection{Data Preparation Pipeline}
\label{ssec:data_preparation_pipeline}
Age estimation models require only a specific region of an image, specifically the person's face, as input, rather than the entire image. However, the influence of this selection process on the model's performance is not apriori known. Should the model be presented with a tight crop of the face or the entire head? Additionally, facial images can differ in terms of scale and resolution since they originate from various sources and as such need to be resized to a uniform resolution. In this section, we examine the impact of the aforementioned data preparation pipeline on the performance of age estimation models. We demonstrate that changes in the preprocessing have a more substantial impact on performance than changes to the loss function.

\begin{figure}
  \centering
  \begin{subfigure}{0.32\linewidth}
    \includegraphics[width=0.86\linewidth]{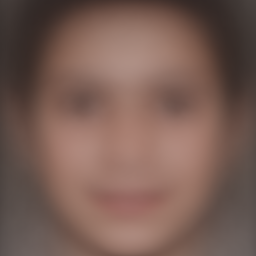}
    \caption{Crop.}
    \label{fig:alignment_mean_face:cropped}
  \end{subfigure}
  \hfill
  \begin{subfigure}{0.32\linewidth}
    \includegraphics[width=0.86\linewidth]{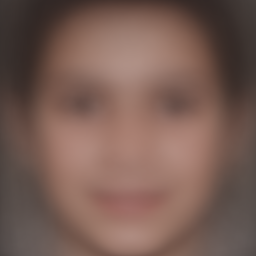}
    \caption{Rotation.}
    \label{fig:alignment_mean_face:rotated}
  \end{subfigure}
  \hfill
  \begin{subfigure}{0.32\linewidth}
    \includegraphics[width=0.86\linewidth]{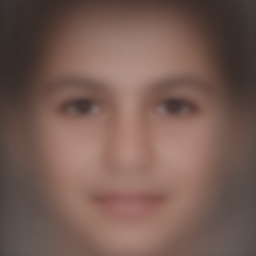}
    \caption{Rot., Trans., Scale.}
    \label{fig:alignment_mean_face:aligned}
  \end{subfigure}
  \caption{Comparison of different alignment methods using the average face from the FG-NET dataset.}
  \label{fig:alignment_mean_face}
\end{figure}

 \paragraph{Facial Alignment}
Numerous studies lack an explanation of their facial alignment procedure. Others merely mention the utilization of facial landmarks. To assess whether a standardized alignment is needed for a fair comparison of multiple methods, we adopt three distinct alignment procedures and evaluate their effect on model performance. Firstly, we \begin{enumerate*}[label=(\roman*),before=\unskip{ }, itemjoin={{, }}, itemjoin*={{, }}]
    \item perform no alignment and employ the bounding box proposed by the facial detection model \cite{RetinaFace} as the simplest approach. The bounding box sides are parallel to the axes of the image. Secondly
    \item we utilize the proposed bounding box but rotate it to horizontally align the eyes. Lastly
    \item we use an alignment procedure, which normalizes the rotation, positioning, and scale. For details, refer to the implementation.
\end{enumerate*} A visual representation of these facial alignment methods is depicted in \cref{fig:alignment_mean_face}. The performance of models trained using the various alignment procedures is presented in \cref{tab:alignment}. When working with pre-aligned datasets like AFAD, we observe that procedure (iii) does not yield significant improvements compared to the simpler variants (i) or (ii). Similar results are obtained on datasets collected under standardized conditions, such as the MORPH dataset. However, when dealing with in-the-wild datasets like AgeDB and CLAP2016, we find that alignment (iii) leads to noticeable improvements over the simpler methods. Interestingly, on the UTKFace dataset, which also contains in-the-wild images, approach (ii) of solely rotating the proposed bounding boxes achieves the best outcomes. However, the disparities among the various alignment procedures are not substantial. We therefore argue that any facial alignment technique that effectively normalizes the position, rotation, and scale of the faces would yield comparable results.

\begin{figure}
  \centering
  \begin{subfigure}{0.32\linewidth}
    \includegraphics[width=0.86\linewidth]{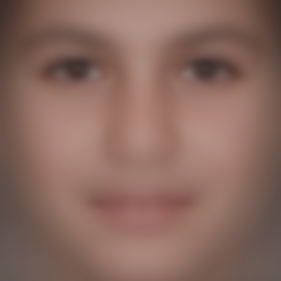}
    \caption{Eyes \& Mouth.}
    \label{fig:scale_mean_face:small}
  \end{subfigure}
  \hfill
  \begin{subfigure}{0.32\linewidth}
    \includegraphics[width=0.86\linewidth]{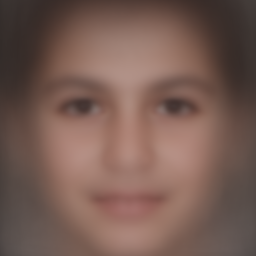}
    \caption{Chin \& Forehead.}
    \label{fig:scale_mean_face:mid}
  \end{subfigure}
  \hfill
  \begin{subfigure}{0.32\linewidth}
    \includegraphics[width=0.86\linewidth]{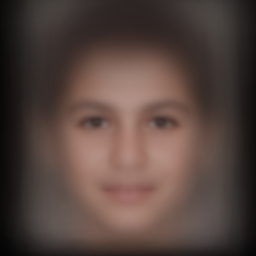}
    \caption{Head.}
    \label{fig:scale_mean_face:large}
  \end{subfigure}
  \caption{Comparison of different facial coverage levels using the average face from the FG-NET dataset.}
  \label{fig:scale_mean_face}
\end{figure}

\paragraph{Facial Coverage}
While facial alignment defines the positioning, orientation, and scale of facial landmarks, the extent to which the face is visible in an image also needs to be specified. We refer to this notion as \textit{facial coverage}. It measures how much of the face is shown in an image and can range from minimal coverage, where only the eyes and mouth are visible, to complete coverage, where the entire head is visible. Determining the optimal compromise between complete facial coverage and minimal coverage is not immediately clear. Complete facial coverage provides a comprehensive view of the face, allowing age estimation algorithms to consider a broader range of facial cues. On the other hand, partial coverage may help reduce overfitting by eliminating irrelevant facial cues and features with high variance. For a visual demonstration of various facial coverage levels, refer to \cref{fig:scale_mean_face}. The concept of facial coverage has received limited attention in age estimation literature. Consequently, the extent of facial coverage utilized in previous studies can only be inferred from the images presented in those works. For instance, Berg et al. \cite{LabelDiversity} seemingly employ minimal coverage, showing slightly more than just the mouth and eyes. The majority of other works \cite{OR-CNN, MovingWindowRegression, DLDL, DLDL-V2, CORAL, Unimodal} tend to adopt partial coverage, where a significant portion of the face, including the chin and forehead, is visible, but not the entire head and hair. In the works of Pan \etal \cite{MeanVariance}, Rothe \etal \cite{DEX}, and Zhang \etal \cite{MegaAge}, the entire head is shown.

The performance of models trained with the different coverage levels is presented in \cref{tab:facial_coverage}.  Generally, complete facial coverage, which includes the entire head in the model input, yields the best results across the majority of datasets. However, specifically for AFAD dataset and the MORPH dataset, partial coverage performs better. It is important to note that the AFAD dataset contains preprocessed images that do not capture the entire head. Consequently, using complete facial coverage with this dataset results in the presence of black bars and a decrease in the effective pixel resolution of the face. It is then to be expected that increased facial coverage yields inferior results. The smallest coverage, limited to the facial region up to the eyes and mouth, consistently performs the worst. With sufficient pixel resolution, the full facial coverage performs the best. % for in-the-wild datasets where the entirety of a person's head can be shown.

\paragraph{Input Resolution}
To investigate the influence of input resolution on age estimation, we performed experiments using multiple resolutions on all datasets: specifically, \mbox{$256\times256$}, \mbox{$128\times128$}, and \mbox{$64\times64$} pixels. The results are presented in \cref{tab:resolution}. Our findings indicate that an increase in image resolution consistently results in improved model performance across all datasets. Hence, the best performance was achieved with a resolution of $256\times256$ pixels. 

In the literature, one can find resolutions ranging from \mbox{$60\times60$} to \mbox{$256\times256$} pixels, where newer works tend to use larger resolution images. As the resolution increase can directly be observed to improve the results; and the resolutions increased with years; it is difficult to say whether newly proposed methods are better overall, or whether they perform better due to using higher resolution images.

 \paragraph{Input Transform}
Finally, we examined the input transformation proposed by Lin \etal \cite{tanh}, which involves converting a face image into a tanh-polar representation. This approach has shown large performance improvements in face semantic segmentation. Lin \etal then modified the network for age estimation, reporting impressive results \cite{FP-AGE}. We explored the potential benefits of applying this transformation for age estimation. However, our findings indicate that the transformation does not improve the results compared to the baseline, as shown in \cref{tab:tanh}. Therefore, we conclude that the improved age estimation performance observed by Lin \etal \cite{FP-AGE} does not arise from the use of a different representation, but rather from pre-training on semantic segmentation or their model architecture.


\begin{table}
\centering
{\small{
\renewcommand{\arraystretch}{0.85}
\begin{tabular}{cccc}
\toprule
\headercell{\\Dataset} & \multicolumn{3}{c@{}}{Alignment}\\
\cmidrule(l){2-4}
& Crop & Rotation & Rot. + Trans. + Scale
\\ 
\midrule
\multicolumn{1}{l}{AgeDB} & 5.93 & 5.92 & \bf5.84\\
\multicolumn{1}{l}{AFAD} & 3.12 & \bf3.11 & \bf3.11\\
\multicolumn{1}{l}{CACD2000} & 4.01 & \bf4.00 & \bf4.00\\
\multicolumn{1}{l}{CLAP2016} & 4.68 & 4.57 & \bf4.49 \\
\multicolumn{1}{l}{MORPH} & 2.81 & \bf2.78 & 2.79\\
\multicolumn{1}{l}{UTKFace} & 4.49 & \bf4.42 & 4.44\\
\bottomrule
\end{tabular}
}}
\caption{MAE $\downarrow$ of ResNet-50 models with different facial alignment. The models were pre-trained on IMDB-WIKI.}
\renewcommand{\arraystretch}{1.0}
\label{tab:alignment}
\end{table}
\begin{table}
\centering
{\small{
\renewcommand{\arraystretch}{0.85}
\begin{tabular}{cccc}
\toprule
\headercell{\\Dataset} & \multicolumn{3}{c@{}}{Facial Coverage}\\
\cmidrule(l){2-4}
& Eyes \& Mouth & Chin \& Forehead & Head
\\ 
\midrule
\multicolumn{1}{l}{AgeDB} & 6.06 & 5.84 & \bf5.81\\
\multicolumn{1}{l}{AFAD} & 3.17 & \bf3.11 & 3.14\\
\multicolumn{1}{l}{CACD2000} & 4.02 & 4.00 & \bf3.96\\
\multicolumn{1}{l}{CLAP2016} & 5.06 & \bf4.49 & \bf4.49\\
\multicolumn{1}{l}{MORPH} & 2.88 & \bf2.79 & 2.81\\
\multicolumn{1}{l}{UTKFace} & 4.63 & 4.44 & \bf4.38\\
\bottomrule
\end{tabular}
}}
\caption{MAE $\downarrow$ of ResNet-50 models with different facial coverages. The models were pre-trained on IMDB-WIKI.}
\renewcommand{\arraystretch}{1.0}
\label{tab:facial_coverage}
\end{table}
\begin{table}
\centering
{\small{
\renewcommand{\arraystretch}{0.85}
\begin{tabular}{cccc}
\toprule
\headercell{\\Dataset} & \multicolumn{3}{c@{}}{Image Resolution}\\
\cmidrule(l){2-4}
& $64\times64$ & $128\times128$ & $256\times256$
\\ 
\midrule
\multicolumn{1}{l}{AgeDB} & 8.43 & 6.90 & \bf5.81\\
\multicolumn{1}{l}{AFAD} & 3.36 & 3.25 & \bf3.14\\
\multicolumn{1}{l}{CACD2000} & 5.01 & 4.55 & \bf3.96\\
\multicolumn{1}{l}{CLAP2016} & 11.34 & 5.90 & \bf4.49\\
\multicolumn{1}{l}{MORPH} & 3.33 & 3.07 & \bf2.81\\
\multicolumn{1}{l}{UTKFace} & 5.83 & 4.81 & \bf4.38\\
\bottomrule
\end{tabular}
}}
\caption{MAE $\downarrow$ of ResNet-50 models with different image resolutions. The models were pre-trained on IMDB-WIKI.}
\renewcommand{\arraystretch}{1.0}
\label{tab:resolution}
\end{table}
\begin{table}
\centering
{\small{
\renewcommand{\arraystretch}{0.85}
\begin{tabular}{ccc}
\toprule
\headercell{\\Dataset} & \multicolumn{2}{c@{}}{Transform}\\
\cmidrule(l){2-3}
& No Transform & RoI Tanh-polar \cite{tanh}
\\ 
\midrule
\multicolumn{1}{l}{AgeDB} & \bf{5.81} & 5.93\\
\multicolumn{1}{l}{AFAD} & \bf{3.14} & 3.15\\
\multicolumn{1}{l}{CACD2000} & \bf{3.96} & 4.07\\
\multicolumn{1}{l}{CLAP2016} & \bf{4.49} & 4.71\\
\multicolumn{1}{l}{MORPH} & 2.81 & \bf{2.80}\\
\multicolumn{1}{l}{UTKFace} & \bf{4.38} & 4.39\\
\bottomrule
\end{tabular}
}}
\caption{MAE $\downarrow$ of ResNet-50 models with different input transformations. The models were pre-trained on IMDB-WIKI \cite{IMDB}.}
\renewcommand{\arraystretch}{1.0}
\label{tab:tanh}
\end{table}

\begin{table}[tb]
\centering
{\small{ \renewcommand{\arraystretch}{0.8}
\begin{tabular}{lcccccc}
\toprule
\headercell{\\Dataset} & \multicolumn{4}{c@{}}{Backbone}\\
\cmidrule(l){2-5}
& \multicolumn{1}{l}{\footnotesize ResNet-50} & \multicolumn{1}{l}{\footnotesize Eff.Net-B4} & \multicolumn{1}{l}{\footnotesize ViT-B-16} & \multicolumn{1}{l}{\footnotesize VGG-16} \\
\cmidrule(l){1-5}
\rotatebox[origin=c]{0}{AgeDB} & 5.81 & \bf5.76 & 9.07 & 6.02\\
\rotatebox[origin=c]{0}{AFAD} & \bf3.14 & 3.20 & 4.04 & 3.22\\
\rotatebox[origin=c]{0}{CACD2000} & 3.96 & 4.00 & 6.22 & \bf3.92\\
\rotatebox[origin=c]{0}{CLAP2016} & 4.49 & \bf4.06 & 8.55 & 4.65\\
\rotatebox[origin=c]{0}{MORPH} & \bf2.81 & 2.87 & 4.35 & 2.88\\
\rotatebox[origin=c]{0}{UTKFace} & 4.38 & \bf4.23 & 6.88 & 4.64\\
\bottomrule
\end{tabular}
}}
\caption{Intra-dataset MAE $\downarrow$ with different backbone architectures. The models were pre-trained on IMDB-WIKI \cite{IMDB}.}
\renewcommand{\arraystretch}{1.0}
\label{tab:model_architecture}
\end{table}

%\begin{table}\centering{\small{\begin{tabular}{cccc}\toprule\headercell{\\Dataset} & \multicolumn{3}{c@{}}{Method}\\\cmidrule(l){2-4}& Cross-Ent. & Mean-Var. \cite{MeanVariance} & Unimod. \cite{Unimodal}\\ \midrule\multicolumn{1}{l}{AgeDB} & 5.81 & 5.75 & 6.17\\\multicolumn{1}{l}{AFAD} & 3.14 & 3.18 & 3.31\\\multicolumn{1}{l}{CACD2000} & 3.96 & 4.06 & 4.23\\\multicolumn{1}{l}{CLAP2016} & 4.49 & 4.30 & 5.25\\\multicolumn{1}{l}{MORPH} & 2.81 & 2.83 & 2.91\\\multicolumn{1}{l}{UTKFace} & 4.38 & 4.49 & 4.66\\\bottomrule\end{tabular}}}\caption{MAE $\downarrow$ of ResNet-50 models.}\label{tab:facial_coverage}\end{table}

\subsection{FaRL Backbone}
We observed that adjustments to the decision layer and loss function have minimal impact on the final model performance. Conversely, large performance disparities arise when modifying other components of the prediction pipeline. Notably, the pretraining data appear to be the most influential factor. Based on this insight, we opt against creating a specialized loss function to enhance the age estimation system. Instead, we leverage the FaRL backbone by Zheng et al. \cite{FaRL}, utilizing a ViT-B-16 \cite{VisionTransformer} model. The FaRL model is trained through a combination of (i) contrastive loss on image-text pairs and (ii) prediction of masked image patches. Training takes place on an extensive collection of facial images (50 million) from the image-text pair LAION dataset \cite{LAION}. We retain the feature representation extracted by FaRL without altering the model's weights. Our decision to use FaRL is driven solely by the extensive amount of pre-training data it incorporates, rather than specific characteristics of the backbone. Different image encoders could be trained in the same manner. However, due to the costs associated with training such models, we have chosen to use the available FaRL ViT-B-16 backbone. We employ a simple multilayer perceptron (MLP) over the FaRL-extracted features, consisting of 2 layers with 512 neurons each, followed by ReLU activation. Cross-entropy serves as the chosen loss function. For each downstream dataset, we pretrain the MLP on IMDB-WIKI or initialize it to random weights. We choose the preferred option based on validation loss on the downstream dataset. As previously, we replace the final layer before fine-tuning on downstream datasets.

This straightforward modification outperformed all other models on AgeDB, CLAP2016, and UTKFace datasets. It also achieved superior results on AFAD, matched the performance of other models on CACD2000, but demonstrated worse performance on MORPH. Applying the FN test revealed statistically significant improvements of this model over others in both intra-dataset and cross-dataset evaluations, see \cref{tab:large_table}. We attribute the poor performance of FaRL on MORPH to the fact that the distributions of images in LAION \cite{LAION} and MORPH \cite{MORPH} are vastly different. As we do not finetune the feature representation of FaRL \cite{FaRL}, it is possible that the representation learned on LAION is superior on the other datasets but deficient on MORPH.

We do not claim this model to be the ultimate solution, but the results achieved with the FaRL backbone along with our public implementation offer a robust and straightforward baseline for a comparison with future methods.

\section{Discussion and Conclusions}
In this paper, we aimed to establish a fair comparison framework for evaluating various approaches for age estimation. We conducted a comprehensive analysis on seven different datasets, namely AgeDB \cite{AgeDB}, AFAD \cite{OR-CNN}, CACD2000 \cite{CACD}, CLAP2016 \cite{CLAP}, FG-NET \cite{FG-NET}, \mbox{MORPH \cite{MORPH}}, and UTKFace \cite{UTKFace}, comparing the models based on their Mean Absolute Error (MAE). To determine if any method outperformed the others, we employed the Friedman test and the Nemenyi critical difference test. When pre-training the models on a large dataset, we did not observe any statistically significant improvement by using the specialized loss functions designed for age estimation. With random model initialization, we observed some improvement over the baseline cross-entropy on small datasets. Specifically, for \mbox{Mean-Variance loss \cite{MeanVariance}}, OR-CNN \cite{OR-CNN}, and DLDL \cite{DLDL}. These improvements can be attributed to implicit regularization provided by these methods. 

Previously published results reported continuous performance improvements over time (as depicted in \cref{fig:year_vs_mae}). Our findings challenge these claims. We argue that the reported improvements can be attributed to either the random data splitting strategy or hyperparameter tuning to achieve the best test set performance. Our analysis of the data preparation pipeline revealed that factors such as the extent of facial coverage or input resolution exert a more significant impact on the results than the choice of the age estimation specific loss function. Guided by these findings, we use the FaRL \cite{FaRL} model as a backbone for age estimation and demonstrated its effectiveness. In summary:
\begin{itemize}

\item We show that existing evaluation practices in age estimation do not provide a consistent comparison of the state-of-the-art methods. We define a proper evaluation protocol which addresses the issue.

\item We show that improvements in age estimation results over recent years can not be attributed to the specialized loss functions introduced in \cite{OR-CNN, DLDL, DLDL-V2, SoftLabels, MeanVariance, Unimodal}, as is claimed in the published literature.

\item Using the insight gained from analyzing different components of the age estimation pipeline, we construct a prediction model with the FaRL \cite{FaRL} backbone and demonstrate its effectiveness.

\item To facilitate reproducibility and simple future comparisons, we have made our implementation framework and the exact data splits publicly available.
\end{itemize}

\section*{Acknowledgment}
This research was supported by the Grant Agency of the Czech Technical University in Prague, grant No. SGS23/176/OHK3/3T/13 and by the Grant agency of the Ministry of Interior Czech Republic, project FACIS grant. No. VJ02010041

\input{cross-dataset-table-colsep}

\clearpage
\clearpage
{
    \small
    \bibliographystyle{ieeenat_fullname}
    \bibliography{main}
}

% WARNING: do not forget to delete the supplementary pages from your submission 
\input{X_suppl}

%\input{X_rebuttal}

\end{document}

%% file: preamble.tex
\definecolor{review1}{rgb}{0.89,0.10,0.10}
\definecolor{review2}{rgb}{0.21,0.5,0.72}
\definecolor{review3}{rgb}{0.3,0.68,0.29}
\definecolor{review4}{rgb}{0.59,0.3,0.64}
\definecolor{review5}{rgb}{1,0.5,0}

\input{macros}

%% file: macros.tex
\definecolor{xpurple}{RGB}{136,46,114}
\definecolor{xblue}{RGB}{25,101,176}
\definecolor{xorange}{RGB}{241,147,45}
\definecolor{xred}{RGB}{220,5,12}
\definecolor{xpink}{RGB}{238, 51, 119}

\definecolor{colorq1}{RGB}{238, 119, 51}
\definecolor{colorq2}{RGB}{0, 119, 187}
\definecolor{colorq3}{RGB}{51, 187, 238}
\definecolor{colorq4}{RGB}{238, 51, 119}

\newcommand\headercell[1]{%
   \smash[b]{\begin{tabular}[t]{@{}c@{}} #1 \end{tabular}}}

\newcommand{\todo}[1]{}
\renewcommand{\todo}[1]{{\color{red} TODO: {#1}}}

\newlength{\dhatheight}

%% file: vf_proposal_eval_protocol.tex
We identified two trivial yet persistent issues that prevent a reliable comparison of age estimation methods. In this section, we address  the initial challenge concerning consistent data partitioning. We provide clear guidelines for the evaluation protocol to ensure replicable and fair assessments.
Specifically, the protocol should establish a reproducible approach for defining the data used in both (i) training and (ii) performance evaluation. When specifying the training data, one needs to state whether the training dataset is the sole source of information, or if the model was \textit{pretrained} with additional data. Additionally, the evaluation can be subdivided based on the data used for \textit{model evaluation} into \textit{intra-dataset}, and \textit{cross-dataset} results. 
%Secondly, one should discern two possible goals when evaluating an age estimation model's performance. Either the goal is to \begin{enumerate*}[label=(\roman*),before=\unskip{ }, itemjoin={{, }}, itemjoin*={{, or to }}]
%\item obtain the \textit{best possible holistic age estimation system}
%\item develop a novel \textit{component} of the existing age estimation pipeline and show its effectiveness. 
%\end{enumerate*} 
We describe how to evaluate models in these settings below.

%In either case, the data used for training needs to be described in detail and the specific data splits must be made public. 

%\textit{Intra-dataset} performance of a model is obtained by \begin{enumerate*}[label=(\roman*),before=\unskip{ }, itemjoin={{, }}, itemjoin*={{, and }}]
%\item splitting a dataset into training, validation, and test parts
%\item training the model using the training and validation parts
%\item measuring the model's performance on the test part
%\end{enumerate*}. 
%\textit{Cross-dataset} performance of a model is obtained by measuring its performance on the entirety of a dataset that was not used for training.

\paragraph{Intra-dataset performance}
To evaluate intra-dataset performance, a single dataset is used for both training and evaluation of the age estimation system. In this case one should \begin{enumerate*}[label=(\roman*),before=\unskip{ }, itemjoin={{, }}, itemjoin*={{, then }}]
\item randomly split the dataset into \textit{subject-exclusive} training, validation, and test set~\footnote{The generated training, validation and test sets will usualy be a partition of the dataset, however, in any case their intersection must be empty.}
\item train the model on the training set
\item measure the model's performance on the validation set
\item possibly revert back to step (ii) and train the model again
\item evaluate the model's performance on the test set
\item publish the results on the test set along with a detailed description of the system components and the data used.
\end{enumerate*} If the dataset consists of limited number of examples, it is possible to create multiple splits of the data into training, validation and test set through step (i). Following this, steps (ii) through (v) are iterated $n$ times, where $n$ is the number of generated splits. It is advisable to present the average test performance along with its standard deviation when reporting the results. 

\paragraph{Cross-dataset performance}
To evaluate cross-dataset performance, the data split in step (i) of the aforementioned evaluation process is generated from a collection of multiple datasets, ensuring that the complete chosen dataset must be employed entirely for evaluation, effectively constituting the designated test set. The remaning steps of the evaluation procedute remain unaltered.

Regardless of the scenario, whether it is intra-dataset or cross-dataset, each system needs to be evaluated against the test data only once, and the results published. All prior model development and hyperparameter tuning must be based solely on the results on the validation set. Furthermore, it should be indicated whether the training data are the only source of information used for training, or whether the model was pretrained with additional data. In the letter scenario, a detailed description of the additional data and their utilization should also be provided.

%% file: cross-dataset-table-colsep.tex
\aboverulesep=0ex
\belowrulesep=0ex
\begin{table*}[]
\begin{adjustbox}{width=\linewidth,center}
\setlength{\tabcolsep}{4pt}
\centering
{\large
{
 \renewcommand{\arraystretch}{1.2}
\begin{tabular}{@{}cc|c*{3}{c}|*{3}{c}|*{3}{c}|*{3}{c}|*{3}{c}|*{3}{c}|*{3}{c}@{}}
\toprule
\multirow{54}{*}{\rotatebox[origin=c]{90}{Training Dataset}}&\multicolumn{1}{c}{} &&\multicolumn{21}{c}{Evaluation Dataset}\\
\cmidrule(l){4-24}
\multicolumn{1}{c}{}&\multicolumn{1}{c}{}&&\multicolumn{3}{c}{AgeDB}&\multicolumn{3}{c}{AFAD}&\multicolumn{3}{c}{CACD2000}&\multicolumn{3}{c}{CLAP2016}&\multicolumn{3}{c}{FG-NET}&\multicolumn{3}{c}{MORPH}&\multicolumn{3}{c}{UTKFace}
\\ 
\cmidrule(l){4-6}
\cmidrule(l){7-9}
\cmidrule(l){10-12}
\cmidrule(l){13-15}
\cmidrule(l){16-18}
\cmidrule(l){19-21}
\cmidrule(l){22-24}
\multicolumn{1}{c}{}&\multicolumn{1}{c}{}&\multicolumn{1}{r}{\diagbox[height=0.7cm]{\small{Method\quad}}{\small{Init.}}}&\multicolumn{1}{c}{IMDB}&\multicolumn{1}{c}{Imag.}&\multicolumn{1}{c}{Rand.}&\multicolumn{1}{c}{IMDB}&\multicolumn{1}{c}{Imag.}&\multicolumn{1}{c}{Rand.}&\multicolumn{1}{c}{IMDB}&\multicolumn{1}{c}{Imag.}&\multicolumn{1}{c}{Rand.}&\multicolumn{1}{c}{IMDB}&\multicolumn{1}{c}{Imag.}&\multicolumn{1}{c}{Rand.}&\multicolumn{1}{c}{IMDB}&\multicolumn{1}{c}{Imag.}&\multicolumn{1}{c}{Rand.}&\multicolumn{1}{c}{IMDB}&\multicolumn{1}{c}{Imag.}&\multicolumn{1}{c}{Rand.}&\multicolumn{1}{c}{IMDB}&\multicolumn{1}{c}{Imag.}&\multicolumn{1}{c}{Rand.}
\\ \cmidrule(l){2-24}
&\multirow{9}{*}{\rotatebox[origin=c]{90}{AgeDB}} 
&\multicolumn{1}{l}{Cross-Entropy}&\cellcolor[gray]{0.9}5.81&\cellcolor[gray]{0.9}7.20&\cellcolor[gray]{0.9}7.65&7.83&12.61&14.70&5.90&8.10& 8.73&6.83&10.86&12.41&10.82&16.27&18.87&4.83&6.74&6.93&8.45&11.88&11.82\\
&&\multicolumn{1}{l}{Regression}&\cellcolor[gray]{0.9}6.23&\cellcolor[gray]{0.9}6.54 &\cellcolor[gray]{0.9}7.60&8.05& 12.13 &14.19&6.76& 7.56 &8.74&8.32& 9.95 &12.32&10.56& 13.83 &18.00&5.64& 6.66 &6.94&9.42& 10.42 &12.14\\
&&\multicolumn{1}{l}{OR-CNN \cite{OR-CNN}}&\cellcolor[gray]{0.9}5.78&\cellcolor[gray]{0.9}6.51 &\cellcolor[gray]{0.9}7.52&7.47& 11.86 &13.80&6.05& 7.71 &8.55&6.64& 10.18 &11.74&9.74& 13.82 &17.40&4.73& 6.61 &7.05&8.20& 10.75 &11.41\\
%&&\multicolumn{1}{l}{CORAL \cite{CORAL}}&\\
&&\multicolumn{1}{l}{DLDL \cite{DLDL}}&\cellcolor[gray]{0.9}5.80&\cellcolor[gray]{0.9}6.95 &\cellcolor[gray]{0.9}7.46&7.81& 1.85 &14.92&5.99& 7.96 &8.52&6.51& 10.84 &11.48&9.23& 15.63 &16.95&4.74& 6.34 &6.82&7.97& 11.87 &11.24\\
&&\multicolumn{1}{l}{DLDL-v2 \cite{DLDL-V2}}&\cellcolor[gray]{0.9}5.80&\cellcolor[gray]{0.9}6.87 &\cellcolor[gray]{0.9}7.58&7.61& 12.98 &16.19&5.91& 8.15 &8.61&6.48& 10.88 &12.50&9.91& 15.36 &19.01&4.92& 7.53 &7.20&7.97& 11.61 &12.30\\
&&\multicolumn{1}{l}{SORD \cite{SoftLabels}}& \cellcolor[gray]{0.9}5.81&\cellcolor[gray]{0.9}6.93 &\cellcolor[gray]{0.9}7.58&7.81& 12.80 &15.42&5.96& 7.90 &8.77&6.61& 10.37 &12.22&9.72& 14.76 &17.85&4.76& 6.58 &7.14&8.12& 11.53 &11.60\\
&&\multicolumn{1}{l}{Mean-Var. \cite{MeanVariance}}&\cellcolor[gray]{0.9}5.85&\cellcolor[gray]{0.9}6.69 &\cellcolor[gray]{0.9}7.33&7.26& 12.40 &14.35&6.00& 7.89 &8.35&6.70& 10.50 &11.90&10.55& 14.32 &17.43&4.99& 6.87 &7.33&8.25& 10.77 &11.64\\
&&\multicolumn{1}{l}{Unimodal \cite{Unimodal}}&\cellcolor[gray]{0.9}5.90&\cellcolor[gray]{0.9}7.11 &\cellcolor[gray]{0.9}15.49&8.37& 13.11 &20.87&6.22& 8.24 &16.11&6.73& 11.14 &21.31&10.15& 16.13 &32.77&4.84& 6.78 &17.42&8.23& 11.86 &23.09\\
\cmidrule(l){3-24} &&\multicolumn{1}{l}{FaRL + MLP} &\cellcolor[gray]{0.9}5.64 & \cellcolor[gray]{0.9}- & \cellcolor[gray]{0.9}- & 7.82 & - & - &7.41 & - & - & 9.32 & - & - &9.15& - & - &4.73& - & - &9.92& - &  - \\ \cmidrule(l){2-24}
&\multirow{9}{*}{\rotatebox[origin=c]{90}{AFAD}} 
&\multicolumn{1}{l}{Cross-Entropy}&15.70&17.31&18.05&\cellcolor[gray]{0.9}3.14&\cellcolor[gray]{0.9}3.17&\cellcolor[gray]{0.9}3.32&9.54&11.18&11.21&8.96&10.23&10.32&10.92&11.38&11.96&6.80&6.83&8.19&12.10&13.07&13.29\\
&&\multicolumn{1}{l}{Regression}&13.67& 15.91 &17.21&\cellcolor[gray]{0.9}3.17&\cellcolor[gray]{0.9}3.16 &\cellcolor[gray]{0.9}3.30&8.72& 10.51 &10.72&8.33& 9.91 &10.02&11.20& 11.89 &12.35&6.27& 7.34 &7.99&11.23& 12.83 &12.96\\
&&\multicolumn{1}{l}{OR-CNN \cite{OR-CNN}}&12.08& 15.65 &16.72&\cellcolor[gray]{0.9}3.16&\cellcolor[gray]{0.9}3.17 &\cellcolor[gray]{0.9}3.28&8.87& 11.05 &10.89&7.85& 9.73 &9.92&10.63& 11.94 &12.58&6.68& 6.85 &7.81&10.50& 12.43 &12.74\\
%&&\multicolumn{1}{l}{CORAL \cite{CORAL}}&\\
&&\multicolumn{1}{l}{DLDL \cite{DLDL}}&14.12& 15.70 &17.21&\cellcolor[gray]{0.9}3.14&\cellcolor[gray]{0.9}3.16&\cellcolor[gray]{0.9}3.25&9.40& 10.70 &11.06&8.68& 9.54 &9.98&11.31& 11.64 &12.07&7.04& 6.75 &7.82&11.52& 12.43 &12.82\\
&&\multicolumn{1}{l}{DLDL-v2 \cite{DLDL-V2}}&13.90& 16.33 &17.78&\cellcolor[gray]{0.9}3.15&\cellcolor[gray]{0.9}3.17 &\cellcolor[gray]{0.9}3.28&9.46& 10.68 &11.02&8.60& 9.76 &10.32&10.83& 11.81 &12.64&6.92& 6.79 &7.94&11.29& 12.61 &13.18\\
&&\multicolumn{1}{l}{SORD \cite{SoftLabels}}&14.30& 16.08 &17.49&\cellcolor[gray]{0.9}3.14&\cellcolor[gray]{0.9}3.15 &\cellcolor[gray]{0.9}3.24&9.45& 10.70 &11.09&8.64& 9.79 &10.10&11.21& 11.63 &12.19&6.87& 6.82 &7.93&11.59& 12.79 &13.10\\
&&\multicolumn{1}{l}{Mean-Var. \cite{MeanVariance}}&12.54& 15.07 &16.68&\cellcolor[gray]{0.9}3.16&\cellcolor[gray]{0.9}3.16 &\cellcolor[gray]{0.9}3.26&8.98& 10.33 &10.75&7.93& 9.33 &9.78&10.96& 12.24 &12.43&6.61& 6.76 &7.88&10.57& 12.00 &12.62\\
&&\multicolumn{1}{l}{Unimodal \cite{Unimodal}}&13.99& 15.89 &20.97&\cellcolor[gray]{0.9}3.20&\cellcolor[gray]{0.9}3.24&\cellcolor[gray]{0.9}9.30&9.23& 10.68 &14.56&8.64& 9.79 &14.51&11.31& 11.83 &18.29&7.07& 7.32 &12.53&11.26& 12.33 &17.47\\
\cmidrule(l){3-24} &&\multicolumn{1}{l}{FaRL + MLP} &16.41 & - & - & \cellcolor[gray]{0.9}3.12 & \cellcolor[gray]{0.9}- & \cellcolor[gray]{0.9}- &10.95 & - & - & 8.57 & - & - &12.24& - & - &6.62& - & - &11.64& - &  - \\ \cmidrule(l){2-24}
&\multirow{9}{*}{\rotatebox[origin=c]{90}{CACD2000}} 
&\multicolumn{1}{l}{Cross-Entropy}&9.66&11.84&10.60&10.70&8.50&13.08&\cellcolor[gray]{0.9}3.96&\cellcolor[gray]{0.9}4.59&\cellcolor[gray]{0.9}4.89&8.42&8.64&10.51&17.45&23.64&20.86&7.21&12.20&10.39&11.16&11.38&12.61\\
&&\multicolumn{1}{l}{Regression}&10.91& 10.44 &10.76&10.23& 7.23 &11.66&\cellcolor[gray]{0.9}4.06&\cellcolor[gray]{0.9}4.52 &\cellcolor[gray]{0.9}4.83&8.84& 7.75 &9.98&17.55& 19.50 &19.60&8.61& 8.81 &11.79&11.34& 10.38 &11.78\\
&&\multicolumn{1}{l}{OR-CNN \cite{OR-CNN}}&10.43& 11.02 &11.85&9.66& 9.48 &12.17&\cellcolor[gray]{0.9}4.01&\cellcolor[gray]{0.9}4.60&\cellcolor[gray]{0.9}4.74&8.57& 8.85 &10.29&18.47& 24.32 &20.85&7.52& 10.04 &11.05&11.17& 12.30 &12.27\\
%&&\multicolumn{1}{l}{CORAL \cite{CORAL}}&\\
&&\multicolumn{1}{l}{DLDL \cite{DLDL}}&9.84& 10.79 &11.28&10.09& 9.30 &13.20&\cellcolor[gray]{0.9}3.96&\cellcolor[gray]{0.9}4.42&\cellcolor[gray]{0.9}4.76&8.39& 8.49 &9.99&18.38& 18.99 &21.52&7.27& 9.16 &11.01&11.19& 11.94 &12.27\\
&&\multicolumn{1}{l}{DLDL-v2 \cite{DLDL-V2}}&9.90& 12.31 &11.20&8.03& 11.50 &11.51&\cellcolor[gray]{0.9}3.96&\cellcolor[gray]{0.9}4.57&\cellcolor[gray]{0.9}4.69&7.67& 8.88 &9.43&18.11& 22.89 &19.02&7.20& 13.46 &9.73&10.52& 12.32 &11.47\\
&&\multicolumn{1}{l}{SORD \cite{SoftLabels}}&9.77& 10.90 &11.04&10.35& 9.55 &11.95&\cellcolor[gray]{0.9}3.96&\cellcolor[gray]{0.9}4.42 &\cellcolor[gray]{0.9}4.70&8.38& 8.51 &9.89&18.05& 20.84 &21.73&7.23& 8.98 &11.59&11.18& 12.06 &12.22\\
&&\multicolumn{1}{l}{Mean-Var. \cite{MeanVariance}}&10.81& 11.42 &10.83&9.71& 10.82 &11.49&\cellcolor[gray]{0.9}4.07&\cellcolor[gray]{0.9} 4.60 &\cellcolor[gray]{0.9}4.78&8.88& 9.20 &10.08&20.48& 22.68 &20.14&8.14& 12.59 &11.72&11.74& 12.29 &12.23\\
&&\multicolumn{1}{l}{Unimodal \cite{Unimodal}}&10.46& 11.04 &46.26&10.63& 9.85 &25.74&\cellcolor[gray]{0.9}4.10&\cellcolor[gray]{0.9} 4.73 &\cellcolor[gray]{0.9}37.41&9.19& 8.92 &30.96&19.37& 19.75 &15.84&8.94& 11.64 &32.63&11.89& 11.75 &32.98\\
\cmidrule(l){3-24} &&\multicolumn{1}{l}{FaRL + MLP} &11.32 & - & - & 9.08 & - & - &\cellcolor[gray]{0.9}3.96 & \cellcolor[gray]{0.9}- & \cellcolor[gray]{0.9}- & 8.57 & - & - &19.63& - & - &6.56& - & - &11.27& - &  - \\ \cmidrule(l){2-24}
&\multirow{9}{*}{\rotatebox[origin=c]{90}{CLAP2016}} 
&\multicolumn{1}{l}{Cross-Entropy}&7.35&10.15&12.26&5.41&7.03&5.34&6.65&8.11&9.11&\cellcolor[gray]{0.9}4.49&\cellcolor[gray]{0.9}5.96&\cellcolor[gray]{0.9}8.73&5.92&9.28&12.02&4.96&6.61&6.90&5.74&7.21&8.58\\
&&\multicolumn{1}{l}{Regression}&7.51& 8.52 &11.74&6.07& 5.19 &5.95&6.86& 7.24 &9.45&\cellcolor[gray]{0.9}4.65&\cellcolor[gray]{0.9} 4.77 &\cellcolor[gray]{0.9}7.89&4.85& 6.31 &10.14&5.09& 5.49 &8.83&6.02& 5.93 &8.66\\
&&\multicolumn{1}{l}{OR-CNN \cite{OR-CNN}}&6.83& 8.74 &11.24&5.83& 5.92 &5.44&6.73& 7.25 &8.65&\cellcolor[gray]{0.9}4.13&\cellcolor[gray]{0.9}4.60 &\cellcolor[gray]{0.9}7.38&5.09& 6.47 &9.22&4.92& 5.78 &6.52&5.43& 5.95 &7.68\\
%&&\multicolumn{1}{l}{CORAL \cite{CORAL}}&\\
&&\multicolumn{1}{l}{DLDL \cite{DLDL}}&7.20& 9.33 &11.39&5.57& 6.90 &5.85&6.85& 7.64 &9.26&\cellcolor[gray]{0.9}4.18& \cellcolor[gray]{0.9}5.10 &\cellcolor[gray]{0.9}7.39&5.26& 7.44 &9.18&4.89& 5.92 &6.52&5.51& 6.37 &7.87\\
&&\multicolumn{1}{l}{DLDL-v2 \cite{DLDL-V2}}&7.14& 9.42 &12.36&5.47& 5.95 &6.45&6.69& 7.99 &9.34&\cellcolor[gray]{0.9}4.23& \cellcolor[gray]{0.9}4.87 &\cellcolor[gray]{0.9}8.52&5.22& 7.04 &8.75&4.85& 6.04 &7.29&5.53& 6.12 &8.23\\
&&\multicolumn{1}{l}{SORD \cite{SoftLabels}}&7.19& 9.60 &12.16&5.47& 7.74 &6.62&6.63& 8.09 &9.66&\cellcolor[gray]{0.9}4.27& \cellcolor[gray]{0.9}5.34 &\cellcolor[gray]{0.9}7.81&5.59& 7.77 &7.62&4.92& 6.01 &6.62&5.48& 6.46 &8.08\\
&&\multicolumn{1}{l}{Mean-Var. \cite{MeanVariance}}&7.08& 9.16 &12.58&5.18& 6.30 &5.38&6.64& 7.37 &9.94&\cellcolor[gray]{0.9}4.28& \cellcolor[gray]{0.9}4.87 &\cellcolor[gray]{0.9}7.95&5.45& 6.69 &11.14&4.96& 7.38 &7.49&5.52& 6.16 &8.65\\
&&\multicolumn{1}{l}{Unimodal \cite{Unimodal}}&7.01& 9.77 &20.71&5.58& 6.10 &5.54&6.47& 8.20 &13.08&\cellcolor[gray]{0.9}4.17& \cellcolor[gray]{0.9}5.39 &\cellcolor[gray]{0.9}13.83&5.13& 6.39 &15.13&4.80& 6.05 &10.02&5.44& 6.67 &15.27\\
\cmidrule(l){3-24} &&\multicolumn{1}{l}{FaRL + MLP} &7.50 & - & - & 4.34 & - & - &6.57 & - & - & \cellcolor[gray]{0.9}3.38 & \cellcolor[gray]{0.9}- & \cellcolor[gray]{0.9}- &4.95& - & - &4.47& - & - &4.85& - &  - \\ \cmidrule(l){2-24}
&\multirow{9}{*}{\rotatebox[origin=c]{90}{MORPH}} 
&\multicolumn{1}{l}{Cross-Entropy}&9.66&11.73&12.63&6.69&7.78&10.36&8.53&10.83&10.11&6.90&8.96&10.64&9.45&11.96&15.38&\cellcolor[gray]{0.9}2.81&\cellcolor[gray]{0.9}2.96&\cellcolor[gray]{0.9}3.01&8.97&10.81&11.92\\
&&\multicolumn{1}{l}{Regression}&10.48& 12.99 &12.56&6.60& 6.65 &10.66&9.82& 11.47 &9.68&7.83& 9.27 &10.67&9.24& 10.13 &16.69&\cellcolor[gray]{0.9}2.83& \cellcolor[gray]{0.9}2.74 &\cellcolor[gray]{0.9}2.97&9.40& 10.97 &12.06\\
&&\multicolumn{1}{l}{OR-CNN \cite{OR-CNN}}&9.35& 11.65 &12.82&6.78& 7.78 &11.81&8.39& 11.34 &10.23&6.84& 8.73 &11.05&9.58& 11.09 &17.47&\cellcolor[gray]{0.9}2.83& \cellcolor[gray]{0.9}2.85 &\cellcolor[gray]{0.9}2.99&8.82& 10.37 &12.06\\
%&&\multicolumn{1}{l}{CORAL \cite{CORAL}}&\\
&&\multicolumn{1}{l}{DLDL \cite{DLDL}}&9.41& 12.00 &12.66&6.58& 7.78 &11.76&8.58& 11.92 &10.10&6.85& 9.26 &11.15&9.44& 11.43 &16.94&\cellcolor[gray]{0.9}2.81& \cellcolor[gray]{0.9}2.92 &\cellcolor[gray]{0.9}2.98&8.80& 10.81 &12.46\\
&&\multicolumn{1}{l}{DLDL-v2 \cite{DLDL-V2}}&9.79& 11.49 &12.68&6.60& 8.22 &12.45&8.79& 10.98 &9.81&6.98& 8.98 &11.22&9.52& 11.63 &17.57&\cellcolor[gray]{0.9}2.82& \cellcolor[gray]{0.9}2.93 &\cellcolor[gray]{0.9}3.00&8.97& 10.70 &12.47\\
&&\multicolumn{1}{l}{SORD \cite{SoftLabels}}&9.48& 11.84 &12.73&6.54& 7.91 &11.19&8.73& 11.18 &10.13&6.84& 8.99 &10.72&9.34& 11.08 &15.90&\cellcolor[gray]{0.9}2.81& \cellcolor[gray]{0.9}2.91 &\cellcolor[gray]{0.9}2.99&8.83& 10.85 &11.97\\
&&\multicolumn{1}{l}{Mean-Var. \cite{MeanVariance}}&9.70& 11.62 &12.93&6.68& 7.81 &10.41&8.65& 10.59 &10.11&7.03& 8.80 &10.56&9.51& 11.45 &15.81&\cellcolor[gray]{0.9}2.83& \cellcolor[gray]{0.9}2.89 &\cellcolor[gray]{0.9}2.95&8.94& 10.59 &11.95\\
&&\multicolumn{1}{l}{Unimodal \cite{Unimodal}}&9.93& 12.31 &17.44&6.63& 7.04 &8.18&8.68& 10.11 &12.03&7.19& 8.95 &12.38&9.80& 12.17 &17.83&\cellcolor[gray]{0.9}2.78& \cellcolor[gray]{0.9}2.90 &\cellcolor[gray]{0.9}8.66&9.07& 10.75 &15.45\\
\cmidrule(l){3-24} &&\multicolumn{1}{l}{FaRL + MLP} &8.40 & - & - & 4.67 & - & - &7.45 & - & - & 6.21 & - & - &9.28& - & - &\cellcolor[gray]{0.9}3.04& \cellcolor[gray]{0.9}- & \cellcolor[gray]{0.9}- &8.93& - &  - \\ \cmidrule(l){2-24}
&\multirow{9}{*}{\rotatebox[origin=c]{90}{UTKFace}} 
&\multicolumn{1}{l}{Cross-Entropy}&6.61&8.88&9.58&5.51&6.42&6.75&6.56&9.10&8.98&4.82&7.34&7.50&4.78&6.62&7.64&5.09&6.61&7.35&\cellcolor[gray]{0.9}4.38&\cellcolor[gray]{0.9}4.75&\cellcolor[gray]{0.9}5.32\\
&&\multicolumn{1}{l}{Regression}&7.01& 7.79 &8.83&5.96& 6.26 &6.43&6.77& 7.87 &8.61&5.24& 5.93 &6.67&4.41& 5.07 &7.27&5.41& 5.95 &6.71&\cellcolor[gray]{0.9}4.72& \cellcolor[gray]{0.9}4.53 &\cellcolor[gray]{0.9}5.34\\
&&\multicolumn{1}{l}{OR-CNN \cite{OR-CNN}}&6.71& 8.29 &8.75&5.56& 6.74 &6.52&6.61& 8.89 &8.37&4.95& 6.79 &6.70&4.54& 5.71 &6.55&5.26& 6.07 &6.76&\cellcolor[gray]{0.9}4.40& \cellcolor[gray]{0.9}4.43 &\cellcolor[gray]{0.9}5.15\\
%&&\multicolumn{1}{l}{CORAL \cite{CORAL}}&\\
&&\multicolumn{1}{l}{DLDL \cite{DLDL}}&6.65& 8.60 &9.00&5.42& 6.68 &6.19&6.52& 9.01 &8.84&4.81& 7.19 &7.46&4.85& 5.87 &7.28&5.16& 6.25 &7.03&\cellcolor[gray]{0.9}4.39& \cellcolor[gray]{0.9}4.66 &\cellcolor[gray]{0.9}5.30\\
&&\multicolumn{1}{l}{DLDL-v2 \cite{DLDL-V2}}&6.79& 8.43 &8.91&5.45& 6.65 &5.82&6.61& 9.39 &8.50&4.98& 7.27 &6.85&4.79& 5.93 &7.44&5.46& 6.24 &6.79&\cellcolor[gray]{0.9}4.42& \cellcolor[gray]{0.9}4.60 &\cellcolor[gray]{0.9}5.19\\
&&\multicolumn{1}{l}{SORD \cite{SoftLabels}}&6.61& 8.96 &9.11&5.42& 7.18 &6.32&6.52& 9.42 &8.69&4.82& 7.87 &7.18&4.83& 6.15 &7.55&5.14& 6.36 &7.28&\cellcolor[gray]{0.9}4.36& \cellcolor[gray]{0.9}4.68 &\cellcolor[gray]{0.9}5.25\\
&&\multicolumn{1}{l}{Mean-Var. \cite{MeanVariance}}&6.79& 8.36 &8.53&5.41& 6.54 &6.32&6.55& 8.55 &8.32&5.04& 6.81 &6.32&5.05& 6.30 &6.90&5.37& 6.15 &6.39&\cellcolor[gray]{0.9}4.42& \cellcolor[gray]{0.9}4.57 &\cellcolor[gray]{0.9}5.05\\
&&\multicolumn{1}{l}{Unimodal \cite{Unimodal}}&6.68& 8.66 &22.42&5.35& 7.68 &16.64&6.58& 9.28 &17.17&4.86& 7.60 &18.83&4.55& 6.25 &22.98&5.22& 5.96 &16.44&\cellcolor[gray]{0.9}4.47& \cellcolor[gray]{0.9}4.78 &\cellcolor[gray]{0.9}21.01\\
\cmidrule(l){3-24} &&\multicolumn{1}{l}{FaRL + MLP} &7.16 & - & - & 4.69 & - & - &6.93 & - & - & 4.02 & - & - &5.07& - & - &4.76& - & - &\cellcolor[gray]{0.9}3.87& \cellcolor[gray]{0.9}- &  \cellcolor[gray]{0.9}- \\
\bottomrule
\end{tabular}
}}
\end{adjustbox}
\caption{Intra-dataset and cross-dataset Mean Absolute Error (MAE) $\downarrow$ of ResNet-50 models. Results marked as \textit{Initialization: IMDB} are of models that are initialized to ImageNet weights, then trained with Cross-Entropy on IMDB-WIKI \cite{IMDB} and then finetuned on the downstream dataset. \textit{Imag.} signifies initialization to weights pre-trained on ImageNet. \textit{Rand.} denotes random initialization.}
\renewcommand{\arraystretch}{1}
\label{tab:large_table}
\end{table*}

%% file: X_suppl.tex
\clearpage
\onecolumn
\setcounter{page}{1}
\maketitlesupplementary
\appendix

\section{Literature Survey}
To thoroughly survey the CVPR and ICCV, we adhered to the following procedure. We systematically searched for all papers presented at the CVPR or ICCV conferences from 2013 onwards that encompassed keywords such as "age", "aging", "face", "facial", and "ordinal" in their titles. We excluded papers that primarily focused on face detection, recognition, or editing, as our specific focus was on age estimation. However, we retained papers that addressed the learning of facial representations (such as unsupervised pre-training and clustering), as they could evaluate the quality of their representations on the age estimation task.

Subsequently, we meticulously reviewed the remaining papers to determine whether they centered around age estimation. We documented the datasets employed by these papers and their data partitioning strategies. Additionally, we extended our survey to include age estimation literature from other conferences that were referenced by the aforementioned papers.

Our estimates, based on the surveyed literature, indicate that approximately $70\%$ of the papers that attempt to improve age estimation follow this approach. Only a minority of these papers adequately ablate the impact of the proposed modifications. Most of the remaining papers suggest modifications to the training procedure, backbone, or aspects of the data pipeline.

\section{Compared Methods}
\label{ssec:age_estimation_methods}
This paper compares various recent age estimation methods utilizing feedforward neural networks which receive an image $x\in\mathcal{X}$ and output an age estimate $y\in\mathcal{Y}$. We focus solely on age estimation methods that modify the standard classification approach by changing the last few layers of the neural network or the loss function. Although this may appear restrictive, it is essential to note that a majority of the methods proposed in the field fall into this category. By comparing methods that modify only a small part of the network, we aim to ensure a fair evaluation, as the remaining setup can be kept identical. Some recent methods, such as Moving Window Regression proposed by \mbox{Shin \etal \cite{MovingWindowRegression}}, were therefore omitted from this study.

Traditionally, age estimation relied on classification and regression-based approaches. However, these methods often overlook the inherent ordinal nature of age. In multi-class classification, misclassifications are treated equally, even though some age predictions may be more accurate than others. On the other hand, regression approaches can predict nonsensical and even negative age values. Ordinal regression has therefore emerged as a well-motivated approach to address these limitations. Unlike classification, where the labels merely represent categories, ordinal regression utilizes labels that provide sufficient information to order the objects. Below, we provide a concise overview of recent age estimation and ordinal regression approaches.

\paragraph{Classification}
The conventional classification approach still remains popular in the literature. For instance, Rothe \etal \cite{DEX} achieved victory in the ChaLearn LAP 2015 challenge on apparent age estimation \cite{CLAP2015_Results} with a model that employed cross-entropy to learn the posterior age distribution.

 \paragraph{Extended Binary Classification}
Niu \etal \cite{OR-CNN} (OR-CNN) follow the approach proposed by Li and Lin \cite{EBC} and transform the ordinal regression task into multiple binary classification sub-problems. For each age value $y_k \in \mathcal{Y}$, they construct a binary classifier to predict whether the true age $y \in \mathcal{Y}$ of a sample $x \in \mathcal{X}$ is larger than $y_k$. Cao \etal \cite{CORAL} (CORAL) modify this approach by restricting the hypothesis class such that the binary classifier predictions are consistent, i.e., the predicted probabilities satisfy ${p(y > y_k | x) \geq p(y > y_{k+1} | x);\, \forall k}$.

 \paragraph{Fixed Distribution Learning}
Gao \etal \cite{DLDL} (DLDL) approach the task as multi-class classification. However, they encode the label distribution as a normal distribution centered at the true label. D\'iaz and Marathe \cite{SoftLabels} (SORD) approach the task similarly, but encode the label distribution as a double exponential distribution centered at the true label. In a follow up to their work \cite{DLDL}, Gao \etal \cite{DLDL-V2} (DLDL-v2) propose to also minimize the difference between \begin{enumerate*}[label=(\roman*),before=\unskip{ }, itemjoin={{, }}, itemjoin*={{, and }}]
    \item the true label $y \in \mathcal{Y}$
    \item the expectation $\mathbb{E}_{\hat y\sim f(x)}[\hat y]$ of the model output distribution $f(x)$.
\end{enumerate*}

 \paragraph{Adaptive Distribution Learning}
An approach emerging in recent years is not to model a specific distribution, such as normal or double exponential distribution, but instead, to constrain the model by some statistical measure or a condition. Pan \etal \cite{MeanVariance} (Mean-Variance) approach the task as standard multi-class classification, but design a loss function that \begin{enumerate*}[label=(\roman*),before=\unskip{ }, itemjoin={{, }}, itemjoin*={{, and }}]
\item minimizes the squared difference between the expectation $\mathbb{E}_{\hat y\sim f(x)}[\hat y]$ and the true label $y \in \mathcal{Y}$
\item minimizes the variance $\mathbb{E}_{\bar y\sim f(x)}\left[ (\bar y-\mathbb{E}_{\hat y\sim f(x)}\left[\hat y\right])^2  \right]$ of the model output distribution $f(x)$.
\end{enumerate*}
Similarly, Li \etal \cite{Unimodal} (Unimodal) design a loss function \begin{enumerate*}[label=(\roman*),before=\unskip{ }, itemjoin={{, }}, itemjoin*={{, and }}]
\item which constrains the model to output unimodal distributions
\item concentrates the output distribution around the true label $y \in \mathcal{Y}$.
\end{enumerate*}

\paragraph{Note on Prediction Strategy}
Note that for all methods which model the posterior distribution $p(y|x)$, namely \begin{enumerate*}[label=(\roman*),before=\unskip{ }, itemjoin={{, }}, itemjoin*={{, and }}]
    \item cross-entropy
    \item \mbox{DLDL \cite{DLDL}}
    \item DLDL-v2 \cite{DLDL-V2}
    \item SORD \cite{SoftLabels}
    \item Mean-Variance loss \cite{MeanVariance}
    \item Unimodal loss \cite{Unimodal}
\end{enumerate*}, we use the optimal plugin Bayes predictor for MAE loss, i.e., we predict $\arg\min_y \mathbb{E}_{\hat y\sim f(x)}[|y - \hat y|]$. For regression, we use the absolute error as the loss function.

\section{Additional Comments}

\paragraph{Pre-training}
For some experiments, we pre-train the models on IMDB-WIKI. However, it is important to note that the labels (bounding box, identity, age) in the IMDB-WIKI dataset are known to be noisy. To mitigate this problem, Lin \etal \cite{FP-AGE}, and Franc and \v Cech \cite{EM-CNN} attempted to clean the labels. To assess the quality of these labels, we trained ResNet-50 models on the dataset using the labels proposed by Lin \etal \cite{FP-AGE} and Franc and \v Cech \cite{EM-CNN}, and evaluated the models' performance on the other datasets \cite{AgeDB, CLAP, CACD, MORPH, UTKFace, OR-CNN}. The results are presented in \cref{tab:imdb-clean}. Both models achieved similar results, so the choice of labels between \cite{FP-AGE, EM-CNN} is in our opinion arbitrary. Due to a slightly lower overall Mean Absolute Error (MAE), we decided to use the labels from Franc and \v Cech \cite{EM-CNN} for model pre-training in this paper.

\begin{table}
\centering
{\small{
\renewcommand{\arraystretch}{1.0}
\begin{tabular}{ccc}
\toprule
\headercell{Evaluation\\Dataset} & \multicolumn{2}{c@{}}{Annotations}\\
\cmidrule(l){2-3}
& EM-CNN \cite{EM-CNN} & FP-AGE \cite{FP-AGE}
\\ 
\midrule
\multicolumn{1}{l}{AgeDB} & 6.44 & \bf{6.30} \\
\multicolumn{1}{l}{AFAD} & \bf{6.86} & 7.23 \\
\multicolumn{1}{l}{CACD2000} & \bf{5.81} & 5.90 \\
\multicolumn{1}{l}{CLAP2016} & 6.24 & \bf{5.53} \\
\multicolumn{1}{l}{FG-NET} & 10.32 & \bf{6.09} \\
\multicolumn{1}{l}{MORPH} & \bf{4.94} & 5.30 \\
\multicolumn{1}{l}{UTKFace} & 8.31 & \bf{6.26} \\
\multicolumn{1}{l}{\textit{Overall}} & \bf{6.28} & 6.36 \\
\midrule
\multicolumn{1}{l}{IMDB} & 4.90 & 5.15 \\
\bottomrule
\end{tabular}
}}
\caption{MAE $\downarrow$ of ResNet-50 trained on IMDB-WIKI with clean age labels from (i) EM-CNN \cite{EM-CNN}, and (ii) FP-AGE \cite{FP-AGE}. Results on IMDB-WIKI are not included in the \textit{Overall} result.}
\renewcommand{\arraystretch}{1.0}
\label{tab:imdb-clean}
\end{table}

\paragraph{Comment on Task Uncertainty}
The irreducible Bayes error of the age estimation task is contingent on the specific formulation. When estimating the \textit{real age} from an observation $x_i$, the label $y_i$ is a realization of the distribution $p(y\,|\,x)$, a random variable. The observation $x_i$ does not contain all the necessary information about the person's genetics, lifestyle, etc., and the Bayes error is non-zero. Interestingly, when estimating the \textit{apparent age}, the Bayes error can be 0. Specifically, when the label is defined as $y_i = \mathbb{E}_{\hat y_i \sim p(y\,|\,x)}[{\hat y_i}]$, which is \textit{not} a random variable. Apparent age is by definition the expected annotation provided by observers. Therefore the observation $x_i$ must necessarily contain all the information used to generate the label. The process of dataset collection provides further insight. For AgeDB, AFAD, and CACD2000, the age label is defined as the \textit{year when the photo was taken} minus the \textit{year of birth}. Two individuals born on December 31st and January 1st would thus have different age labels, even though they were born just a day apart. The \textit{year taken} and the \textit{year born} can also be noisy. For instance, AFAD collects data from a social network with a minimum age requirement of 15, without verifying the actual age of the users. CLAP2016, FG-NET, and MORPH use the legal age as the age label. Still, uncertainty arises from the discrete nature of the label. For UTKFace, the labels are estimated by DEX \cite{DEX}, and manually verified. Therefore, DEX should attain zero error on UTKFace.

%\clearpage
%\clearpage
%%%%%%%%% REFERENCES
%{
%    \small
%    \bibliographystyle{ieeenat_fullname}
%    \bibliography{main}
%}